# Hardware-Efficient Deconvolution-Based GAN for Edge Computing

Azzam Alhussain
*College of Engineering and Computer Science*
*University of Central Florida*
Orlando, Florida, USA
mr.azzam@knights.ucf.edu

Mingjie Lin
*College of Engineering and Computer Science*
*University of Central Florida*
Orlando, Florida, USA
Mingjie.Lin@ucf.edu

*Abstract*—Generative Adversarial Networks (GAN) are cutting-edge algorithms for generating new data samples based on the learned data distribution. However, its performance comes at a significant cost in terms of computation and memory requirements. In this paper, we proposed an HW/SW co-design approach for training quantized deconvolution GAN (QDCGAN) implemented on FPGA using a scalable streaming dataflow architecture capable of achieving higher throughput versus resource utilization trade-off. The developed accelerator is based on an efficient deconvolution engine that offers high parallelism with respect to scaling factors for GAN-based edge computing. Furthermore, various precisions, datasets, and network scalability were analyzed for low-power inference on resource-constrained platforms. Lastly, an end-to-end open-source framework is provided for training, implementation, state-space exploration, and scaling the inference using Vivado high-level synthesis for Xilinx SoC-FPGAs, and a comparison testbed with Jetson Nano.

## I. Introduction

Deep Neural Networks (DNNs) achieve state-of-the-art performance in a wide range of applications, including image classification [1], object detection [2], and speech recognition [3]. Recently, a class of DNN called Generative Adversarial Network (GAN) had been frequently used for data generation [4], as individuals and businesses have become more data conscious. The basic GAN architecture is illustrated in Fig. 1, demonstrating an innovative technique for training a model by framing the issue as an unsupervised learning problem with two distinct sub-models: the generator *G* model is trained to generate new samples, while the discriminator *D* model is trained to distinguish between real and fake data. Moreover, its performance has increased dramatically over the last several years, and they are now capable of producing high-quality, visually attractive, and realistic pictures [5]. However, their superior performance comes at the cost of increased computational power, significant energy consumption, and memory needs.

Researchers have recently made numerous efforts to shrink the size of DNN algorithms to enable their deployment on low-power edge devices. For example, they have examined methods to reduce computational costs by direct quantization [6], quantization-aware training [7][8][9], pruning and compression [10], the use of less demanding layers (depth-wise separable convolutions) [11], and the technique of non-arithmetic layers (e.g., ShiftNet [12]).

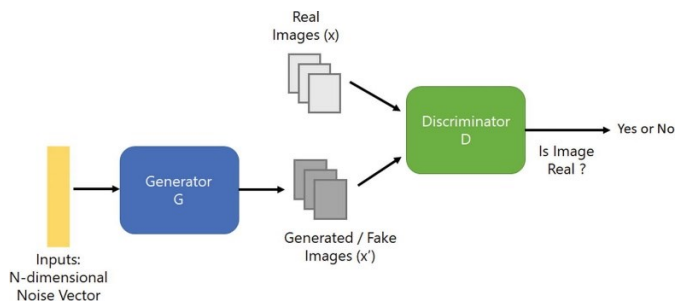

Figure 1. Generative Adversarial Network (picture taken from [13])

Compared to the original DNNs, these methodologies have reduced the computational cost and eased their deployment on several embedded platforms (e.g., NVIDIA Jetson Nano, Avnet's Ultra96, and Raspberry Pi) due to their high performance and low power consumption. They are capable of achieving an acceptable to a high throughput depending on the application. Jetson Nano includes a tiny embedded graphics processing unit (GPU), whereas the Ultra96 platform includes a Xilinx field-programmable gate array (FPGA) with a hard ARM processor system on chip (SoC-FPGA), enabling developers to design custom accelerators for any algorithm. For this adaptability, this research concentrates on SoC-FPGAs implementation. Moreover, FINN by Xilinx [14] is an open-source research framework for accelerating DNNs inference on FPGAs. They provide efficient building blocks for training the quantized neural network (QNN) and developing inference accelerators while only supporting limited DNN layers.

In this paper, FINN was extended to train a quantized deconvolutional GAN (DCGAN) [15], designed a scalable accelerator, and deployed the inference on SoC-FPGAs. We also benchmarked our implementation against NIVIDA Jetson Nano and achieved a superior throughput to power ratio when running the inference. The contributions of this research are as follows:

- Developed a scalable inference accelerator for transpose convolution operation for quantized DCGAN (QDCGAN) on top of FINN.

- Provided a complete open-source framework [16] (training to implementation stack) for investigating the effect of variable bit widths for weights and activations.

- Demonstrated that the weights and activations influence performance measurement, resource utilization, throughput, and the quality of the generated images.

The rest of this paper is organized as follows: Section II provides background on DCGAN and its hardware implementation. Section III discusses the design process of the network architecture and the accelerator mechanism. Section IV describes the experimental tools and presents the results, and Section V concludes the paper.

## II. BACKGROUND

DCGANs are used to generate new images from the learned data distribution. It occasionally incorporates a block called deconvolution or fractionally stride convolution. Deconvolution is the spatial inverse of convolution as it transforms the input to a higher output dimension, as shown in Fig. 2. When applied to photos, it results in an up-sampling [17]. While most open-source efforts focus on Convolution Neural Network (CNN), which employs an ordinary convolution, DCGAN has received less attention regarding the implementation on edge platforms as compared to convolution and CNN. Deconvolution is used in image super-resolution GAN (SRGAN) [18] to map a low-resolution picture to a higher-resolution image. It is specially used in microscopic imaging [19] to obtain high-quality photos and produce results in real-time. An efficient implementation of deconvolution also enables the deployment of super-resolution on low-power edge devices and opens the possibility of having technologies like NVIDIA deep learning super-sampling (DLSS) on low-power edge computing.

The most recent work that is in line with our idea is proposed by Colbert et al.[20], which implemented a deconvolution CNN (DCNN) in the PYNQ-Z2 device. Nevertheless, this research is qualitatively distinct from the past one. Several issues are introduced and solved in this paper. First, it is constrained to a 32-bit fixed-point implementation without experimenting with other bit widths. Second, the accelerator is a non-scalable systolic array, which implies that each layer is executed sequentially by a single-engine, resulting in increased latency. Third, the weight of each layer is streamed from the DDR (there is no on-chip storage for all the network weights), resulting in increased power consumption. Lastly, it does not make the quantized training, accelerator design, and deployment code open-source to enable experimentation with different network sizes and accelerator design choices.

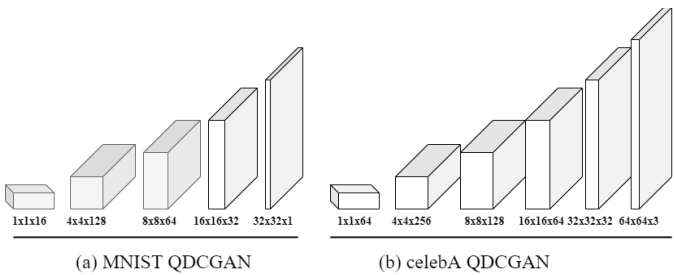

(a) MNIST QDCGAN  (b) celebA QDCGAN

Figure 2. QDCGAN architectures for inference acceleration

## III. DESIGN PROCESS

### A. Network Architecture

Training GAN comes in a zero-sum game between two competing networks; the generator $G$ and the discriminator $D$. $G$ tries to maximize the loss of $D$ by mapping a noise vector to the input space while $D$ objectives to maximize the chance to identify the real distribution of data. Basically, $G$ is trained to fool $D$ as in the equation follows:

$$min_G max_D\ E_{x \sim p_r}[log\ D(x)] + E_{\tilde{x} \sim p_g}[\log(1 - D(\tilde{x}))] \quad (1)$$

Where $Pr$ is the data distribution, and $Pg$ is the model distribution implied by $\tilde{x} = G(z), z \sim p(z)$ (the input $z$ to the generator is sampled from the noise distribution $p$). Since GAN are inherently unstable, Wasserstein GAN loss (WGAN) along with the gradient penalty proposed by Gulrajani et al.[21] are applied to stabilize the training and optimize $G$. The WGAN loss function equation is as follows:

$$min_G max_D\ E_{x \sim p_r}[D(x)] + E_{\tilde{x} \sim p_g}[D(\tilde{x})] \quad (2)$$

Adding the gradient penalty term as follows:

$$L = E_{\tilde{x} \sim p_g}[D(\tilde{x}) - \{E_{x \sim p_r} * D(x)\} + \lambda * \\ E_{\hat{x} \sim P_{\hat{x}}}\{(\|\nabla_{\hat{x}} D(\hat{x})\|_2 - 1)^2\}] \quad (3)$$

$D$ works as a continuous function and is not trained to classify the data. $P\hat{x}$ represent a straight line between data distribution of $Pr$ & $Pg$ with a fixed coefficient $\lambda = 10$ that has been proven to work on various architectures and datasets. Besides, applying the previous critic model eliminates the need for batch normalization [22] in any network as well as calculating different thresholds for every layer and only using the simple thresholds (quantize ReLU), which resulted in better images quality. In contrast, weights and activations were quantized in the forward pass using a hard tanh function followed by an n-bit width quantizer. The following equation is for rounding and clipping the weights:

$$Q(x, n) = Clip\left\{\frac{round(x*2^n)}{2^n}, -1, 1\right\} \quad (4)$$

$x$ is the input, whereas $n$ represents the number of bits. If the magnitude of the input during the forward pass is $1 < x \leq -1$, the straight-through estimator (hard tanh) will pass the gradients. Yet, if the input is outside this range, hard tanh will zero-out the gradients according to the following equation:

$$g_x = \{g_q * 1_{|x| \leq 1}\} \quad (5)$$

Prior equations were applied to improve and stabilize the training of QDCGAN on MNIST and celebA datasets. These publicly available datasets were the most commonly used for new GANs. Besides, the developed QDCGAN architectures shown in Fig. 2 were trained in full-precision weight (W) and activation (A), e.g., W32A32, and then fine-tuned to lower bit-widths (fixed point), e.g., W4A4 and W1A2.

### B. Accelerator Design

The accelerator was built on top of FINN, and it uses the processing element (PE) and the single instruction multiple data (SIMD) scaling factor for each layer separately. Every layer has its engine for execution. The resources utilization versus throughputs is balanced by variable precisions and numbers of PE & SIMD, which define the inputs and outputs parallelism respectively. However, the overall parallelism is calculated by the folding factor (FF) as in the following equation:

$$FF = \left\{\frac{H*W}{PE*SIMD}\right\} \quad (6)$$

$H$ and $W$ represent height and width of the matrix respectively. The smaller the folding factor, the lower the latency to execute the weight matrix. Moreover, the developed accelerator uses a transpose convolution that works by applying certain degrees of expansion and padding of zeroes in-between the input feature map values. The expansion and padding values are determined by the stride and filter size of the deconvolution layer. Using this pre-processing step of learnable up-sampling deconvolution can be implemented with an efficient convolution accelerator. Fig. 3 demonstrates an example of how this method helps in the implementation of GAN-based edge devices. The sliding window generator for deconvolution uses a circular ring buffer which has an efficient mechanism for efficiently maintaining and moving a list of values systematically. It helps to store a small portion of the input feature map channels on-chip that is sufficient to generate a sliding window and improve memory utilization.

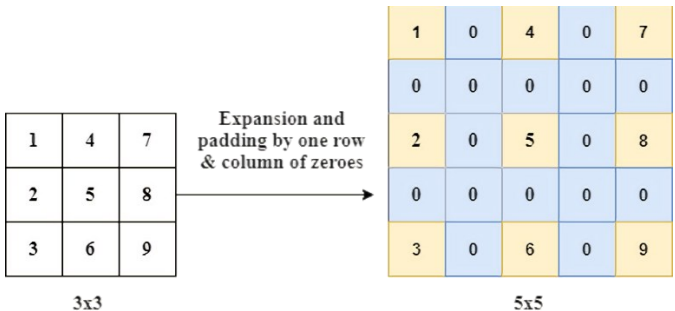

Figure 3. Pre-processing of up-sampling (expansion) for implementing deconvolution operation as a convolution.

Subsequently, the accelerator receive inputs from the processor's DDR, executes the layers, and outputs the results back to the processor. The weights are then stored on-chip (BRAM) and can be accessed concurrently by the partitioned PE elements in parallel, which results in fast execution. After training, the generator network's weights are then extracted, packed according to PE and SIMD of each layer, and saved as an FPGA-readable format. Hence, the proposed accelerator is then introduced in FINN-HLSLIB open-source library [23], synthesized to run at 125 MHz, and implemented on SoC-FPGAs. To this end, the network architecture is customizable for larger platforms and datasets.

*C. Host Code*

Python with PYNQ API [24] is used to write the driver code that communicates with the hardware, load the accelerator bitstreams into the programmable-logic (PL), loads the weights, passes input/output buffers from DDR, and convert the output data into a visual format to the user.

## IV. EXPERIMENTAL SETUP AND RESULTS

*A. Development Environment*

Nvidia Tesla P4 GPU was used to train the QDCGAN. PyTorch was used as a training framework, while Brevitas library from Xilinx [25] was used for quantization aware-training. Brevitas is a flexible quantization library utilized for training quantized DNNs and deconvolution as well. It is also capable of adjusting and controlling the bit width of each layer's weight and activation as well as performing multiply and accumulates (MAC) operations.

On the hardware side, Vivado synthesized the accelerator bitstream and analyzed the design architecture. Additionally, FINN-HLSLIB is one of the development tools utilized and built upon it in this research. It is an open-source library to develop and implement an efficient QNN accelerator using Xilinx high-level synthesis (HLS). PYNQ is another open-source project-based used for Zynq platforms that provide a Python framework and APIs to load the bitstream and run the inference. Fig. 4 shows the steps followed of the training environment and hardware implementation.

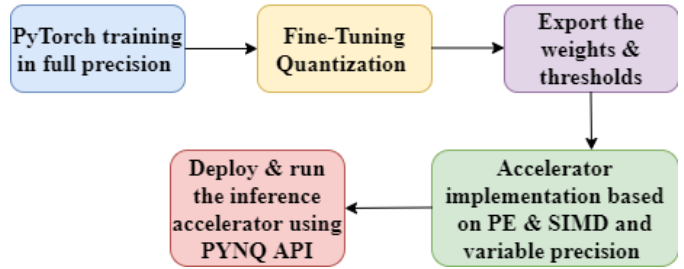

Figure 4. Block diagram of training and hardware environment steps

After implementing and generating the results, the developed accelerator appeared to be very efficient due to the streaming dataflow architecture, on-chip weights storage, and being fully pipelined where each layer has its engine to speed up the processing, reduce the latency, and provide far more performance per watt than the GPU and CPU.

*B. Implementation Results*

Characteristics of hyperparameters during training play a significant role in achieving promising results of QDCGAN, in which trials and errors were the baselines of this research. MNIST and celebA datasets were trained from scratch on full-precision, fine-tuned to the lowest possible bit-width, and achieved aesthetically clear generated images. Since MNIST is a greyscale dataset and has one channel, the lowest possible bit-width utilized were W1A2 and W4A4. Conversely, celebA is a large colored dataset and has three channels, and the lowest possible bit-width utilized was W4A4. Moreover, the Fréchet Inception Distance (FID) [26] was employed to quantify the quality of generated images for various precisions. There is no specific range or scale for measuring the FID. Although, a lower score indicates better-quality images. Hence, the generated images for both datasets are illustrated in Fig. 5 shown different precisions with an associated FID score for each one.

The developed accelerator-based deconvolution proved to be efficient in terms of less resource usage. Table 1 shows the resource utilization for Ultra96 and ZCU104 in runtime weights for both datasets. Moreover, running the inference with runtime configurable weights produces 2.1x-2.5x times higher throughput than the baked-in weights with less than 10% increase in the resource utilization reported. Lastly, the network architecture, scalability, throughput in frame per second (FPS), actual power consumption when running the inference, and the proposed accelerator efficiency (performance per watt) is reported and benchmarked against Jetson Nano in Table 2.

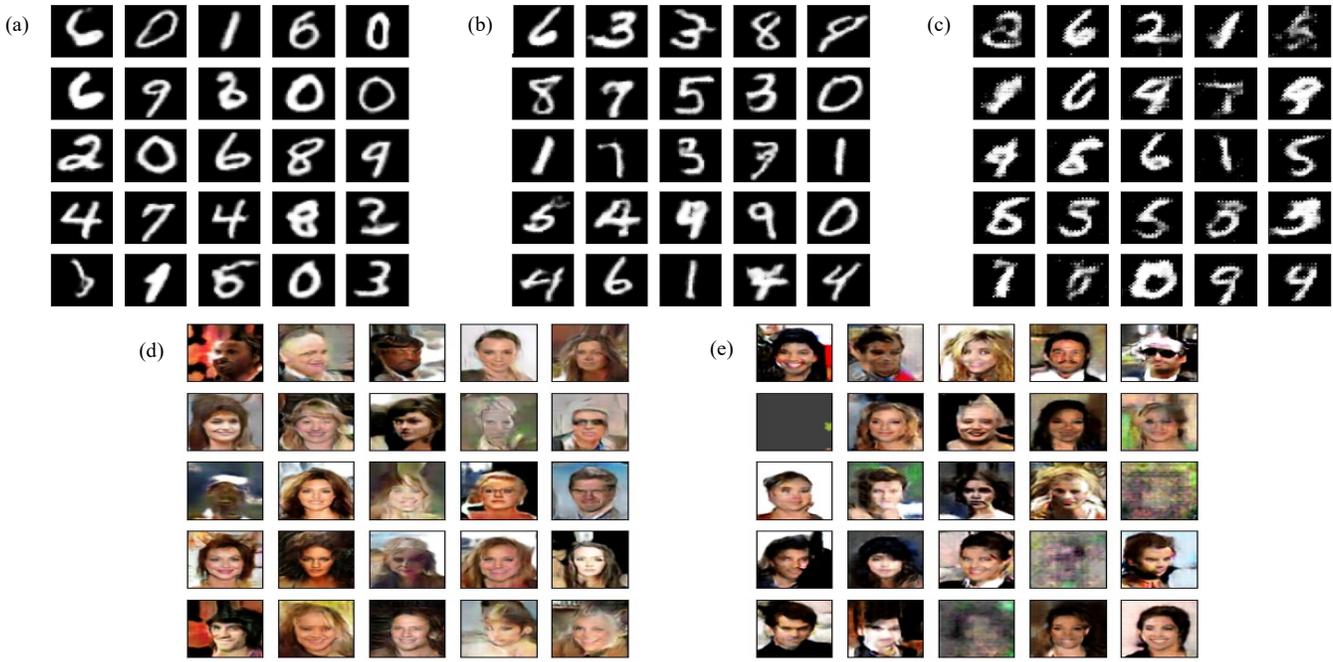

Figure 5. MNIST and celebA were trained from scratch on W32A32 for 100 epochs and resulted in (a) and (d) with an FID score of 49 and 104 respectively. Then, MNIST fine-tuned to a lower bit-width with more training epochs resulted in (b) W4A4 and (c) W1A2 with an FID score of 53 and 126 respectively. Finally, celebA is also fine-tuned to the lowest possible bit-width with more training epochs resulted in (e) W4A4 with an FID score of 129.

Table 1. Resource utilization

| Datasets | MNIST | | celebA | |
|---|---|---|---|---|
| Device | Ultra96 | | Ultra96 | ZCU104 |
| Bit-width / resources | *W1A2* | *W4A4* | *W4A4* | |
| Flip-Flops | 22k (15%) | 141k (26%) | 60k (42%) | 63k (13%) |
| LUTs | 17k (24%) | 70k (46%) | 44k (63%) | 77k (33%) |
| BRAM | 22 (10%) | 48 (22%) | 156 (72%) | 174 (55%) |
| LUTRAM | 1k (4%) | 28k (5%) | 1k (3%) | 1k (1%) |
| DSP Slices | 1 (0.28%) | 1 (0.28%) | 5 (1%) | 5 (0.29 %) |

Table 2. Performance measurement

| Dataset | MNIST | | celebA | | |
|---|---|---|---|---|---|
| Device | Jetson Nano | Ultra96 | Jetson Nano | Ultra96 | ZCU104 |
| Output Channels | [128, 64, 32, 1] | | [256, 128, 64, 32, 3] | | |
| PE | - | [4,8,8,1] | - | [4,8,8,3] | [16,16,16,16,3] |
| SIMD | - | [4,16,16,8] | - | [4,16,16,16,8] | [16,16,16,16,16] |
| FPS | 233-284 | 1802-1813 | 60-76 | 301-312 | 890-904 |
| Power (W) | 4.5-3.3 = 1.2 | 5.5-5.3 = 0.2 | 4.5-3.3 = 1.2 | 5.5-5.3 = 0.2 | 11.9-11.6 = 0.3 |
| FPS/W | 208 | 9K | 58 | 1.5K | 3k |

## V. CONCLUSION AND FUTURE WORK

This paper utilized Xilinx's publicly accessible FINN project to develop a scalable accelerator for QDCGANs. The proposed architecture is based on a ring buffer to generate the sliding windows for deconvolution resulted in higher throughput while using fewer resources. The provided open-source code can contribute to the community to explore and search efficient implementation of SRGAN on low-power FPGAs and assist- deploying NIVIDA DLSS on edge platforms which are considered as a solution for a wide range of medical and microscopic imaging applications.